\documentclass{article}

\usepackage[nonatbib]{neurips_2023}

\usepackage[utf8]{inputenc} 
\usepackage[T1]{fontenc}    
\usepackage{hyperref}       
\usepackage{url}            
\usepackage{booktabs}       
\usepackage{amsfonts}       
\usepackage{nicefrac}       
\usepackage{microtype}      
\usepackage{xcolor}         
\usepackage{graphicx}
\usepackage{amsmath}
\usepackage{mathrsfs}
\usepackage{amsfonts,amssymb}
\usepackage{multirow}
\usepackage{biblatex}
\usepackage{fontawesome}
\usepackage{listings}
\usepackage{color}

\definecolor{dkgreen}{rgb}{0,0.6,0}
\definecolor{gray}{rgb}{0.5,0.5,0.5}
\definecolor{mauve}{rgb}{0.58,0,0.82}

\DeclareMathOperator*{\argmin}{\arg\!\min}

\title{ Heterogeneous Generative Knowledge Distillation with Masked Image Modeling  }

\author{
  Ziming Wang*, Shumin Han\dag, Xiaodi Wang\dag, Jing Hao\dag, Xianbin Cao*, Baochang Zhang* \thanks{Correspondence: bczhang@buaa.edu.cn} \\
  *Beihang University \\
  \dag Baidu.com
}

\begin{document}

\maketitle

\begin{abstract}

 Small CNN-based models usually require transferring knowledge from a large model before they are deployed in computationally resource-limited edge devices. Masked image modeling (MIM) methods achieve great success in various visual tasks but remain largely unexplored in knowledge distillation for heterogeneous deep models. The reason is mainly due to the significant discrepancy between the Transformer-based large model and the CNN-based small network. In this paper, we develop the first Heterogeneous Generative Knowledge Distillation (H-GKD) based on MIM, which can efficiently transfer knowledge from large Transformer models to small CNN-based models in a generative self-supervised fashion. Our method builds a bridge between Transformer-based models and CNNs by training a UNet-style student with sparse convolution, which can effectively mimic the visual representation inferred by a teacher over masked modeling. Our method is a simple yet effective learning paradigm to learn the visual representation and distribution of data from heterogeneous teacher models, which can be pre-trained using advanced generative methods. Extensive experiments show that it adapts well to various models and sizes, consistently achieving state-of-the-art performance in image classification, object detection, and semantic segmentation tasks.  For example, in the Imagenet 1K dataset, H-GKD improves the accuracy of Resnet50 (sparse) from 76.98\% to 80.01\%. 
\end{abstract}

\section{Introduction}
\label{Introduction}

Generative learning has an extraordinary power to learn from unlabeled large-scale data. Inspired by the success in NLP (e.g. BERT  \cite{devlin2018bert}, GPT  \cite{radford2018improving}), masked image modeling methods (MIM) show a solid ability to learn visual representations from the prediction of masked patches. In addition to better learning ability, the burgeoning MIM models pursue deeper and broader structures. However, deploying large networks (such as ResNet101 \cite{he2016deep}) on computationally limited devices is complex. Knowledge distillation (KD)  aims to transfer knowledge from a pre-trained high-capacity teacher network to a lightweight student model, which is beneficial for deploying efficient networks on resource-constrained devices. Pioneering research by Hilton \cite{hinton2015distilling} et al. shows that low-capacity networks (the student model) can often be significantly improved, benefiting from the auxiliary knowledge gained by a set of large models. However, traditional distillation methods have difficulty utilizing large MIM models due to two shortcomings. 

\begin{figure}[htbp]
  \centering
  \includegraphics[width=13cm]{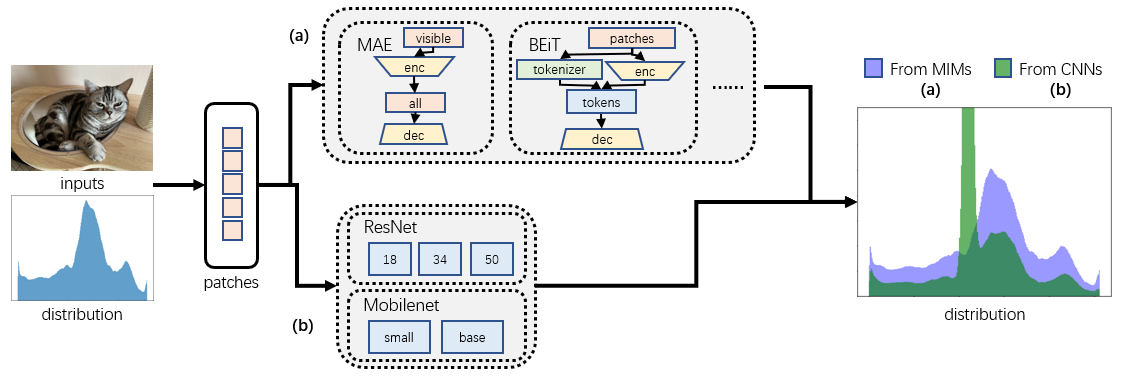}
  \caption{\textbf{The distribution shift.} The above figures show the distribution of pixel intensity histograms  of the student and the teacher. If \textbf{(b)} straightforwardly feed masked images to CNNs (like Resnet \cite{he2016deep}, Mobilenet \cite{howard2017mobilenets}), the data distribution will shift at scale (shown in green). In contrast, \textbf{(a)} Transformer-based generative methods (like MAE \cite{he2022masked}, BEiT \cite{bao2021beit}) have no such side effect due to the ability to process variable-length input (shown in blue).}
  \label{fig:shift}
\end{figure}

First, smaller CNN-based student networks can quickly learn the knowledge (such as logits and features) extracted by multiscale visual processing methods (such as SIFT  \cite{lowe1999object}, Swin  \cite{liu2021swin}) and pyramid networks \cite{lin2017feature} with labeled unmasked images. However, they have difficulty in learning the knowledge from single-scale masked modeling possessed by heterogeneous MIM teacher models (such as CAE  \cite{chen2022context}, MAE  \cite{he2022masked}) and in processing masked images (discussed in Section \ref{Methods:Sparse}). For example, as shown in Fig. \ref{fig:shift}, student models cannot mimic the data distribution of MIM teacher models. One may zero out all masked pixels and directly feeds the masked image to the student. However, compared with the consistent output distribution of teacher models (shown in blue), the result of CNNs leads to a severe data distribution shift (shown in green). Transferring knowledge from the MIM teacher to the CNN student with such an extensive distribution discrepancy can disturb the student's learning pattern. Therefore, directly training a small CNN-based network to distill knowledge using MIM seems not applicable. 

Secondly, self-supervised distillation does not require annotated data compared with traditional supervised KD methods. As a result, it does not learn a probability distribution in some classes or categories \cite{hinton2015distilling}. Therefore,  self-supervised learning from the training data itself becomes particularly important. In addition to a single instance, the relationship between multiple images may demonstrate more about the inner relations of the dataset, which represents the probability distribution. However, traditional distillation strategies mainly transfer features  \cite{chen2022knowledge,passban2020alpkd,chen2021crosslayer} or logits  \cite{hinton2015distilling,zhou2021rethinking,xu2020knowledge} for instance, which is not plausible to learn the inner-relationship of the data itself.

We inject generative learning into knowledge distillation and propose Heterogeneous Generative Knowledge Distillation (H-GKD) as a new learning paradigm to learn data's visual representation and distribution from heterogeneous models. With the analysis above, we find that traditional Convnets with overlapping sliding windows are hard to learn the knowledge from masked images and single-scale outputs. Inspired by sparse convolution \cite{graham2017submanifold}, we design a student model for learning generative visual representations better. Besides being equipped with sparse convolution, the student model also uses a UNet-style  \cite{ronneberger2015unet} to keep the advantages of multiscale structures. Meanwhile, to solve the problem of a learning probability distribution, our idea is first to calculate the distances (similarities) between the input image and other data points, then convert them to distribution. A memory queue is maintained to transfer the distribution from the teacher to the student to achieve this. Specifically, we train a teacher model using generative methods, extracting pre-trained representations and similarity distribution. Subsequently, we train a sparse convolution-based student model using generative methods and align the knowledge from the heterogeneous teacher to the student. This simple structure brings noticeable results:

\begin{itemize}
    \item[-]  Unlike traditional KD methods, generative distillation does not require any labeled data or additional image operations. The student model can efficiently learn the inner relations of the dataset itself through similarity distribution.
    
    \item[-] The teacher model can be pre-trained using  advanced generative methods, such as MAE  \cite{he2022masked}, CAE  \cite{chen2022context}, BEiT  \cite{bao2021beit}, as a result, the student model can learn the visual representation from a heterogeneous teacher via sparse convolution and hierarchical structure. 
\end{itemize}

To demonstrate the effectiveness of the proposed model, a series of downstream tasks are performed on Imagenet 1K \cite{deng2009imagenet} and COCO \cite{lin2014microsoft} datasets, including image classification, object detection, and semantic segmentation. For example, on the Imagenet 1K dataset, H-GKD improved the accuracy of ResNet50 (sparse) from 76.98\% to 80.17\% (a gain of more than 3\%). Compared with state-of-the-art distillation methods, H-GKD achieves a significant improvement in classification, which is greater than 1\%. To summarize, our work provides the following.

\begin{itemize}
    \item[•] We are the first to inject Masked Image Modeling (MIM) methods into Knowledge Distillation. We describe an H-GKD method to transfer knowledge between heterogeneous models in a self-supervised manner.

    \item[•]  We achieve a new learning paradigm to learn the visual representation and distribution of data from heterogeneous teacher models, which can be pre-trained using  advanced generative methods.

    \item[•] With the proposed method, we refresh the state-of-art results achieved by the  SOTA KD methods on several downstream tasks. Compared with various distillation strategies, we conduct extensive experiments under multiple settings on benchmark datasets to verify our effectiveness.
\end{itemize}

\section{Related Work}
\label{Related Work}

\textbf{Self-supervised learning} is a method of learning labels derived from data. Visual self-supervised work has achieved comparable or even better results than pretraining on downstream tasks such as supervised/semi-supervised classification and object detection. The initial application of self-supervised learning was to learn data representations through a pretext task, including rotation prediction \cite{deng2021does}, an exemplar-based method \cite{li2022ckdf}, and a jigsaw \cite{kim2018learning}. Recently, contrastive learning \cite{he2020momentum,chen2020improved,chen2021empirical} has become popular, such as methods that model the correlation \cite{gao2022cross} (or uncorrelated) and dependency between two or more images. Contrastive learning methods are mainly based on data augmentation. Compared to contrastive methods that rely on training strategies, generative methods only rely on unlabeled data. Mask image modeling learns knowledge from images corrupted by masking. The DAE \cite{vincent2008extracting} is a pioneering MIM method that uses masking as a type of noise. The Context Encoder \cite{pathak2016context} uses a convolutional neural network to complete missing areas. Inspired by NLP, transformer-based methods \cite{liu2021swin,dosovitskiy2020image} have become popular, such as iGPT \cite{chen2020generative}, ViT \cite{dosovitskiy2020image}, which studies masked patch prediction, and BEiT \cite{bao2021beit}, which predicts discrete tokens. The MAE \cite{he2022masked} paper studies masking ratio and visible patches, while CAE \cite{chen2022context} focuses on learning capability. Unlike exploring Transformers, SparK \cite{tian2023designing} explores the potential of CNN in MIM methods.

\textbf{Knowledge Distillation (KD)} is  first introduced in the work of Hinton et al. \cite{hinton2015distilling}. Initial research focused on using the softmax output of the teacher network as additional supervision to train the student network. However, the output of high-capacity networks was found to have little difference from the ground-truth labels. Additionally, the softmax output contains less information than the representation from the second-to-last layer due to the classifier layer. Such highly abstract latent knowledge from the last layer loses much encoding information in the hidden layers. Thus, the distillation of features has gained increasing attention, such as attention maps \cite{wei2022b}, activation boundaries \cite{heo2019knowledge}, Gram matrix \cite{li2019layer}, and information from multiple intermediate layers \cite{wu2020skip}. Some studies also focus on feature transfer in intermediate layers \cite{wu2020skip}, such as pairwise relationships \cite{tian2022contrastive} and the distillation of knowledge between layers \cite{chen2021crosslayer}. A reasonable explanation of the success of feature-based transferring intuitively lies in that the learning process of intermediate layers represents the inference direction and inductive bias of the final result. Recent research has focused more on adjusting training strategies (such as data augmentation \cite{gordon2019explaining}, label smoothing \cite{tang2021understanding}), multi-teachers \cite{vongkulbhisal2019unifying,zhao2019highlight,yuan2020reinforced}, and extracting information from different layers \cite{wang2020exclusivity,zagoruyko2017paying}). DIST \cite{huang2022knowledge} uses the Pearson correlation coefficient to narrow the gap between teacher and student. 

\textbf{Self-Supervised Distillation}  methods mainly generate auxiliary labels through image transformations, such as rotation \cite{gidaris2018unsupervised}, and coloring \cite{zhang2016colorful}. Inspired by contrastive learning, recent work (such as SSKD \cite{xu2020knowledge}) explore cross-sample contrast relations through image rotation and augmentation. HSAKD \cite{yang2021hierarchical} learns augmented data distributions between different layers through an auxiliary classifier. CompRess \cite{han2015deep} and SEED \cite{fang2021seed} explore the importance of developing better self-supervised distillation for small models. However, the success of contrastive learning often relies on dataset-specific data augmentation strategies and carefully designed additional training strategies. These strategies are difficult to apply to general tasks. In contrast, generative learning aims to reconstruct the features and information of the sample itself, avoiding dependence on strategies. This paper proposes masked generative modeling with distillation to extract more rich image knowledge. 

\section{H-GKD: Heterogeneous Generative Knowledge Distillation}
\label{Methods}

\subsection{Preliminary on Knowledge Distillation}
\label{Methods:Preliminary}

The training strategy of traditional Supervised Knowledge Distillation \cite{hinton2015distilling} can be generalized as the following formulation:

\[ \hat{\theta} = \argmin_{\theta} \sum_{i}^{N} \mathcal{L}_{dis}(x_i, \delta^T, \theta) + \mathcal{L}_{sup}(x_i, y_i, \theta), \]

where $x_i$ is the input image, $y_i$ is the corresponding ground truth label, $\delta$ is the student parameter set, and $\theta_T$ is the teacher network's output (logits or features). $\mathcal{L}_{sup}$ is the supervised error between the model prediction and the annotation. For example, the classification task is usually a cross-entropy loss for class, while the object detection task usually includes bounding box regression. $\mathcal{L}_{dis}$ denotes the mimic loss of a student towards a pre-trained teacher. For example Kullback-Leibler divergence \cite{van_Erven_2014} can be a loss to measure the prediction that came from softmax. The squared $\mathbf{l} 2$ distance \cite{Dokmanic_2015} also is a loss to  verify the alignment between intermediate feature maps. The effectiveness of the loss above functions has been well proven in supervised learning, but it remains unexplored for the self-supervised heterogeneous settings, which is our different focus.

\subsection{Heterogeneous Generative Knowledge Distillation}
\label{Methods:H-GKD}

Compared with supervised learning, we aim to build a large model using a ready-made MIM algorithm and compress it to a small CNN-based model while preserving the extraordinary power of learning visual concepts. 

\begin{figure}[htbp]
  \centering
  \includegraphics[width=13cm]{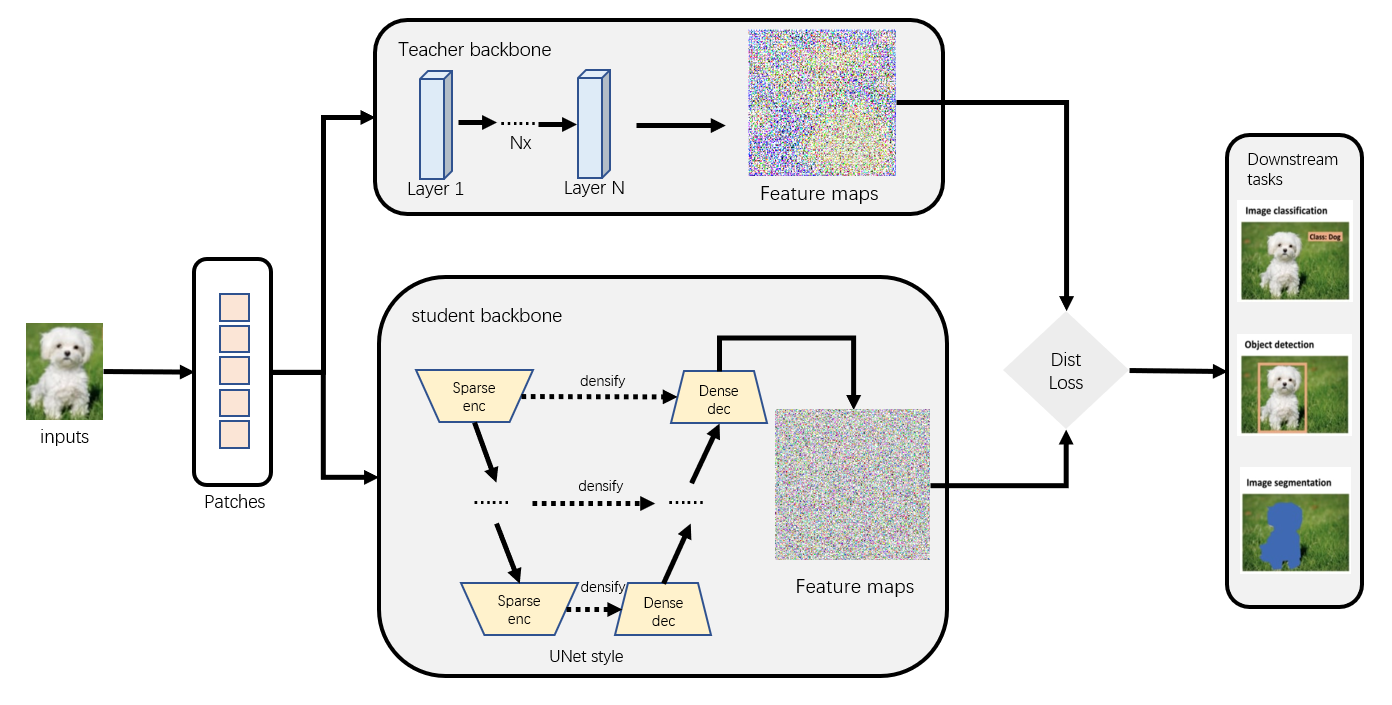}
  \caption{The structure of H-GKD. We first pre-train a MIM model, which could be MAE \cite{he2022masked} or BEiT \cite{bao2021beit} as the teacher model. Then a sparse CNN-based student network would be trained to learn the visual representation of the teacher by minimizing the loss (Section \ref{Methods:Similarity}).}
  \label{fig:pipeline}
\end{figure}

\subsubsection{Ability to Learn Visual Representation}
\label{Methods:Sparse}

\begin{figure}[htbp]
  \centering
  \includegraphics[width=13cm]{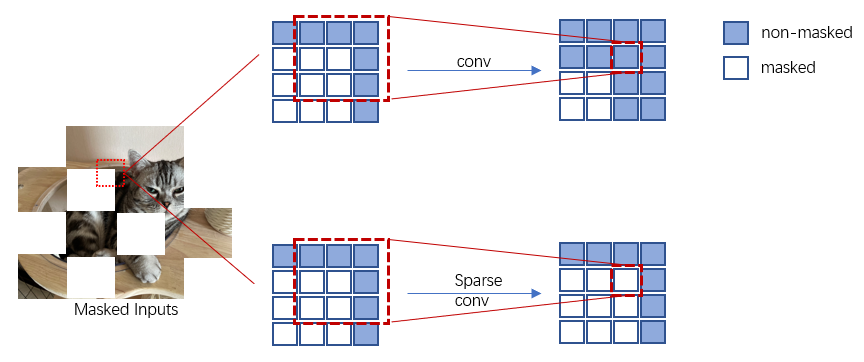}
  \caption{\textbf{Sparse convolution for the masked image.} As the above traditional convolution shows, when applying a kernel centered at a zero pixel, the result could be non-zero if any non-zero pixel exists in the filter. The masked region will be eroded, and the visible one will extend after repeating the convolution. However, sparse convolution would skip the masked pixel, keeping the mask pattern.}
  \label{fig:conv_compare}
\end{figure}

The student's ability to learn from masked images must be verified to learn the visual representation of MIM methods. In masked image modeling, the network is usually fed by images divided into non-overlapping patches. Each patch has a probability of being masked by a mask ratio. Transformer-based masked modeling may directly apply zero to remove the masked patches. Traditional convolution operates on full-valued grids where a sliding kernel calculates each image pixel. Directly eliminating the information of masked patches in CNN-based networks may lead to several problems: 1) when applying traditional convolution on the edge between masked pixels and visible pixels, the results will shrink the masked region as shown in Fig. \ref{fig:conv_compare}, leading to severe mask pattern change after a large number of convolutional blocks. 2) the data distribution may shift as illustrated in Fig. \ref{fig:shift}. To solve these problems, we apply sparse convolution \cite{graham2017submanifold}, which can directly substitute traditional convolution operation in any CNN-based network without backbone readjustment. This approach (as shown in Fig. \ref{fig:conv_compare}) would skip all the masked pixels and only compute at visible points, keeping the mask pattern, thus ensuring no information is leaked. Moreover, the sparse convolution can be naturally translated to the traditional convolution at fine-tuning since dense images are a no-mask case of sparse images.

Hierarchical information is acknowledged as an important part of visual representation. To mimic the teacher's learning ability for hierarchical knowledge, we use the UNet structure. Taking a ResNet model with UNet-design, for example. An image with $\mathbf{H} \times \mathbf{W}$ inputs in the encoder with 4 sparse CNN-based blocks ($E_1, E_2, E_3, and E_4$) to produce 4-scale sparse feature maps ($F_1, F_2, F_3, and F_4$) with resolutions of $\frac{H}{4} \times \frac{W}{4}, \frac{H}{8} \times \frac{W}{8}, \frac{H}{16} \times \frac{W}{16}, and \frac{H}{32} \times \frac{W}{32}$. Subsequently, we utilize  a three-blocks ($D_1, D_2, and D_3$) light decoder with up-sampling layers to process all these hierarchical feature maps, forming a final image representation. $M_1, M_2, M_3, and M_4$ are mask embeddings to fill in all the empty 'holes' in the corresponding sparse feature maps $F_1, F_2, F_3, and F_4$. To erase the width differences between the encoder and the decoder, four projection layers($\phi_1, \phi_2, \phi_3, and \phi_4$) are applied. Therefore, we can get the output ($S_3, S_2, 
and S_1$) of decoder blocks (with shape of $frac{H}{16} \times \frac{W}{16}, \frac{H}{8} \times \frac{W}{8}, \frac{H}{4} \times \frac{W}{4}$):

\[ S_i = D_i(S_{i+1}) + \phi_i(F_i + M_i), (\forall i \in {3,2,1}), \]

where $S_4 = F_4$. The final output of the student model in this example is $S_1$.

\subsubsection{Learning Through Instance Similarity}
\label{Methods:Similarity}

In addition to transfer features, for different architecture families, such as Transformer-based MIM methods and CNN-based networks, transferring the relations between outputs might also be a plausible approach. Inspired by distilling the instance relationships (such as RKD \cite{park2019relational}, MOCO \cite{he2020momentum}), we develop an effective method based on corresponding instance similarity over a memory queue. The output of the teacher is stored in a maintained memory queue. The similarity scores of both the student and the teacher are computed with the memory queue, in which the objective turns to minimize the divergence between the probabilities over similarity scores between the teacher and the student. 

\begin{figure}[htbp]
  \centering
  \includegraphics[width=13cm]{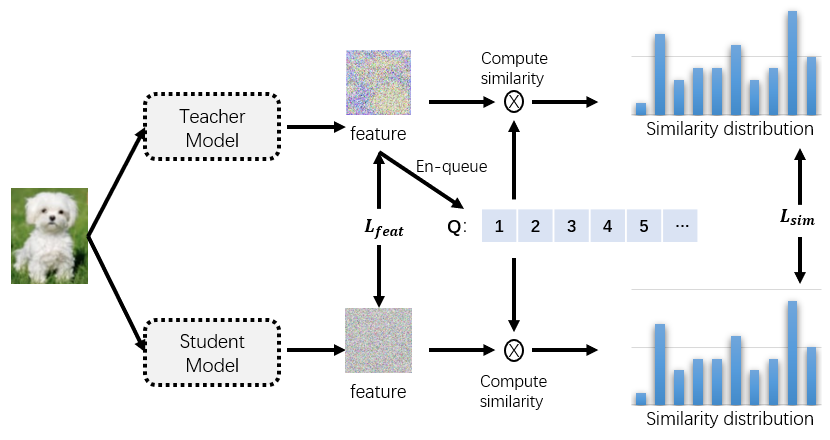}
  \caption{\textbf{The pipeline of our distillation method.} The teacher model is pre-trained and frozen during distillation. The student model is trained by 1) the MSE loss between features of the teacher and the student and 2) the Pearson coefficient loss between the similarity distribution of the teacher and the student.}
  \label{fig:similarity}
\end{figure}

Specifically, as shown in Fig. \ref{fig:similarity}, with the input image $x_i$, the frozen pre-trained teacher $f^{T}$ encodes $x_i$ into an embedding $t^T(x_i) \in \mathbb{R}^N$ with a parameter set $\theta^T$. The goal for the student $f^{S}$ is to learn an embedding $t^S(x_i) \in \mathbb{R}^N$ to mimic the similarity between $t^T(x_i)$ and $t^T(x_j), j \neq i$. A memory queue $\mathbf{Q} \in \mathbb{R}^K$ with FIFO strategy can be used to calculate similarity.\footnote{If the queue length is considerable or even equal to the data length, the results may keep a good performance, but time and space complexity will be bigger.} Since the maintained instances are mostly randomly chosen, the student similarity, computed between a student output and the random-chosen memory queue, is irrelevant to the input instance $x_i$. Thus to avert the disruption of student similarity distribution, we first update the teacher outputs and then calculate the students. That is, the teacher embedding $t^T(x_i)$ en-queue and compute the similarity to form the $\mathbf{Q_{k+1}} = \{q_1, q_2, ... ,q_{k+1}\}$ before the student embedding $t^S(x_i)$. $t^S(x_i)$ does not en-queue and only computes the similarity distribution.

The similarity between the teacher and the student are defined as follows:

\[ \mathbf{P^T_i}(t^T(x_i), \theta | \mathbf{Q_{K}}, x_i) = \frac{exp(t^T(x_i) \cdot t^T(x_j)/\tau)}{\sum_{k=1}^K exp(t^T(x_i) \cdot t^T(x_k)/\tau)},\]

\[\mathbf{P^S_i}(t^S(x_i), \theta | \mathbf{Q_{K}}, x_i) = \frac{exp(t^S(x_i) \cdot t^T(x_j)/\tau)}{\sum_{k=1}^K exp(t^T(x_i) \cdot t^T(x_k)/\tau)},\]

where $\mathbf{P^T_i}(t^T(x_i), \theta | \mathbf{Q_{K}}, x_i)$ and $\mathbf{P^S_i}(t^S(x_i), \theta | \mathbf{Q_{K}}, x_i)$ are similarity computed by teacher feature $t^T(x_i)$, memory $t^T(x_j)$ from the memory queue $\mathbf{Q_{K}}$, and by student feature $t^S(x_i)$, memory $t^T(x_j)$ respectively. $\tau$ denotes a temperature parameter that aligns the value into the same calculated space. For example, the similarity between $t^T(x_i)$ and the current-updated element $t^T(x_{k+1})$ (which is itself) is the largest among the current step, in which this similarity only can be adjusted by $\tau$. Therefore, the loss function of our similarity distribution can be formulated as follows:

\[\mathbf{L}_{sim}(\theta) = -\mathbf{L}_{Pearson}(\mathbf{P^T}(t^T(x), \theta | \mathbf{Q_{K}}, x), \mathbf{P^S}(t^S(x), \theta | \mathbf{Q_{K}}, x)),\]

\[= \frac{\sum (\mathbf{P^T_i}-\hat{\mathbf{P^T}}) (\mathbf{P^S_i}-\hat{\mathbf{P^S}})}{\sqrt{\sum (\mathbf{P^T_i}-\hat{\mathbf{P^T}})^2 \sum (\mathbf{P^S_i}-\hat{\mathbf{P^S}})^2}}, \]

where we maximize the Pearson correlation coefficient (minimize $-\mathbf{L}_{Pearson}$) of the similarity distribution of the teacher and the student. $\hat{\mathbf{P^T}}$ and $\hat{\mathbf{P^S}}$ denote the mean of similarity between the teacher and the student. Moreover, to learn the visual representation of the powerful MIM model, we require the student to mimic the feature output towards each instance. Thus, for the outputs $t^T(x_i)$ and $t^S(x_i)$, a MSE loss are applied:

\[\mathbf{L}_{feat}(\theta) = \sum_i MSE(t^T(x_i), t^S(x_i) | \theta).\]

Overall, our distillation objective can be formulated as follows:

\[ \hat{\theta} = \argmin_{\theta} \mathbf{L}_{sim}(\theta) + \mathbf{L}_{feat}(\theta). \]

\section{Experiments}
\label{Experiments}

\subsection{Experimental Settings}
\label{Experiments:Experimental Settings}

\paragraph{Self-Supervised pre-training settings.} For teacher network, we take CAE \cite{chen2022context} as default. Following the official settings, we use ViT-Large with 24 Transformer blocks as the backbone. The image of 224 $\times$ 224 is partitioned into 14 $\times$ 14 patches with the size of 16 $\times$ 16. We also apply MAE \cite{he2022masked} and BEiT \cite{bao2021beit} as teacher networks. The output size is 1024 $\times$ 196 for one instance. All teachers are pre-trained by original settings (1600 epochs). For the student network, we also apply a self-supervised method to train. By a convolution layer as projection, the student's input and output have the same size as the teachers'. At the distillation training stage, the size of the memory queue is set to 50,000, and the temperature is set to 0.009 after several experiments. We use PyTorch along with LAMB \cite{wightman2021resnet} optimizer with momentum(beta) 0.9, max gradient norm parameter of 5.0, and weight decay of 1e-4 for 100 epochs. The learning rate is set to 8e-4 and updated by 40 warm-up epochs with a cosine decay scheduler. The batch size is 1024 for a single GPU. Moreover, we run our experiments on Tesla V100 with 8 GPUs with 32GB memory.

\paragraph{Downstream Tasks.} We only use unlabeled ImageNet 1K \cite{deng2009imagenet} for all self-supervised and distillation methods, and both ImageNet 1K and MS-COCO \cite{lin2014microsoft} for downstream fine-tuning and evaluation. Linear classification is conducted on ImageNet 1K, and object detection and semantic segmentation are evaluated on MS-COCO. For image classification, we apply the pre-trained student's weights to a backbone that has the same size and depth as the student and append a multi-layer-perceptron layer (two linear layers with a ReLU activation in between) at the end of the fine-tuning model after average pooling. Faster R-CNN \cite{ren2016faster} and Mask R-CNN \cite{he2018mask} are used for object detection and semantic segmentation, respectively, in which after applying the student's weights, we append the Faster R-CNN and Mask R-CNN's neck and head, then fine-tune all the layers of the model.

\paragraph{Other settings.} For fairly comparison, we apply same baseline architecture and pre-training weights on all benchmarks. Table \ref{table:hypers_baseline}
 shows training hyper-parameters for all models, in which CAE-L applies to teacher models and ResNet-50 applies to student models.
 
\begin{table}[htbp]
  \caption{\textbf{Training hyper-parameters and details for baseline models.}}
  \label{table:hypers_baseline}
  \centering
  \resizebox{0.6\linewidth}{!}{
  \begin{tabular}{|l|l|l|l|}
    \toprule
        architecture & CAE-L & ResNet-50 & Sparse R-50\\ 
    \midrule
        Image Size & 224 & 224 & 224 \\ 
    
        Epoch & 1600 & 300 & 300 \\ 
        
        Batch Size  & 2048 & 2048 & 2048 \\ 
        
        Optimizer & lamb \cite{you2020large} & lamb \cite{you2020large} & lamb \cite{you2020large} \\ 
        
        Learning Rate & 0.1 & 0.1 & 0.1 \\ 
        
        LR Scheduler & cosine & cosine & cosine \\ 
        
        Weight Decay & 0.04 & 0.04 & 0.04 \\ 
        
        Random Crop & True & True & True \\ 
        
        Flipping & True & True & True \\ 

        Jittering & True & True & True \\ 

        Top-1 Acc. & 82.60 & 75.81 & 76.98 \\ 
    \bottomrule
  \end{tabular}
  }
\end{table}

\subsection{Evaluation}
\label{Experiments:Evaluation}

\begin{table}[htbp]
  \caption{\textbf{Comparison of distillation methods.} Top-1 accuracy, Bounding Box AP, and Mask AP (\%) are shown in the table. (d-epoch for distillation epoches). All methods uses CAE-L as teacher and Sparse ResNet-50 as student.}
  \label{table:baselines}
  \centering
  \resizebox{0.6\linewidth}{!}{
    \begin{tabular}{|l|l|l|l|l|l|l|l|l|l|}
    \toprule
        method & d-epoch & TOP-1 & TOP-5 & $\rm AP^{bb}$ & $\rm AP^{mk}$ \\ 
    \midrule
        WSLD \cite{zhou2021rethinking} & 100 & 78.23 & 93.01 & 38.25 & 34.24 \\ 
        
        SSKD \cite{xu2020knowledge}  & 100 & 78.72 & 93.05 & 37.64 & 33.10 \\ 
        
        SRRL \cite{yang2021knowledge} & 100 & 79.62 & 94.25 & 37.71 & 33.12 \\ 
        
        SEED \cite{fang2021seed} & 100 & 79.96 & 95.35 & 39.94 & 36.67 \\ 
        
        HSAKD \cite{Yang_2021} & 100 & 79.65 & 94.85 & 38.03 & 34.35 \\ 
        
        DIST \cite{huang2022knowledge} & 100 & \textbf{80.02} & \textbf{95.87} & 38.94 & 35.03 \\ 
        
        Ours & 100 & 80.01 & 95.37 & \textbf{39.74} & \textbf{36.22} \\ 
    \bottomrule
    \end{tabular}
    }
\end{table}

We compare our approach with several state-of-the-art knowledge distillation methods to validate our effectiveness. Table \ref{table:baselines} shows the results of different teacher and student networks. Our method outperforms other distillation methods on all benchmarks. We apply pretrained timm weights \cite{wightman2021resnet} to all teacher and student models for a fair comparison. For example, all teachers who choose ResNet101 as a backbone will load the timm-resnet101 pre-trained weight to facilitate their training performance. All the weights are pre-trained on ImageNet. \textbf{On ImageNet 1K,} our approach achieves the best accuracy over all SOTA methods on ResNet50, and it even catches the accuracy of the supervised model, which is trained for 1600 epochs. \textbf{On MS-COCO,} our results also show the best performance on object detection and semantic segmentation compared to the SOTA distillation methods. Furthermore, Our method also shows an outstanding performance in teaching the student as shown in Table \ref{table:improve_points}. Our model outperforms several SOTA models with different model depths (e.g. ResNet-34/50), especially for deeper models.

\begin{table}[htbp]
  \caption{\textbf{Accuracy Improvement.} We also compare the improvement of accuracy over several SOTA models.}
  \label{table:improve_points}
  \centering
  \resizebox{0.5\linewidth}{!}{
  \begin{tabular}{|l|l|l|l|l|}
    \toprule
        method & student (top-1 acc.) & TOP-1 & $\Delta$ \\ 
    \midrule
        KD \cite{hinton2015distilling} & \multirow{2}{*}{ResNet-34 (73.31)} & 75.37 & +2.06 \\ 
        
        DIST \cite{huang2022knowledge} & ~ & 75.74 & +2.43 \\ 
    \midrule    
        ours & S-ResNet-34 (73.75) & \textbf{76.20} & \textbf{+2.45} \\ 
    \midrule
        KD \cite{hinton2015distilling} & \multirow{4}{*}{ResNet-50 (75.81)} & 78.11 & +2.3 \\ 
        
        RKD \cite{park2019relational} & ~ & 78.59 & +2.78 \\ 
        
        SRRL \cite{yang2021knowledge} & ~ & 78.42 & +2.61 \\ 
        
        DIST \cite{huang2022knowledge} & ~ & 78.73 & +2.92 \\ 
    \midrule    
        ours & S-ResNet-50 (76.98) & \textbf{80.01} & \textbf{+3.03} \\ 
    \bottomrule 
  \end{tabular}
  }
\end{table}

\subsection{Ablation Study}
\label{Ablation Study}

\begin{table}[htbp]
  \caption{\textbf{Stability of our method.} We verify the stability on ImageNet 1K and MS-COCO for downstream tasks by Top-1 accuracy (T-Top-1 for teachers, S-Top-1 for students), Bounding Box AP ($AP^{bb}$), and Segmentation AP ($AP^{mk}$) (\%). For example, the accuracy is stable when distilling from different teachers (all around 82.5\% )}
  \label{table:stability}
  \centering
  \resizebox{0.9\linewidth}{!}{
  \begin{tabular}{|l|l|l|l|l|l|l|l|}
    \toprule
    teacher & student & fine-t ep & det ep & T-Top-1 & S-Top-1 & S-COCO $AP^{bb}$  & S-COCO $AP^{mk}$ \\ 
    \midrule
    CAE-L & s-resnet50 & 100 & 12 & 82.60 & 80.00 & 39.74 & 36.22 \\ 
    
    MAE-L & s-resnet50 & 100 & 12 & 82.43 & 79.93 & 39.98 & 36.54 \\ 
   
    BEiT-L & s-resnet50 & 100 & 12 & 82.51 & 80.01 & 39.55 & 36.01 \\ 
    \midrule
    ~ & s-resnet50 & 100 & 12 & ~ & 76.98 & 38.10 & 34.88 \\ 
    
    CAE-L & s-resnet50 & 100 & 12 & 82.60 & 80.00 & 39.74 & 36.22 \\ 
    
    ~ & s-resnet34 & 100 & 12 & ~ & 72.37 & 34.37 & 31.23 \\ 
    
    CAE-L & s-resnet34 & 100 & 12 & 82.60 & 74.82 & 36.71 & 34.00 \\ 
    
    \bottomrule
  \end{tabular}
  }
\end{table}

\paragraph{Stability.} To verify the effectiveness and stability of our method, we apply different teacher-student pairs for evaluation. Table \ref{table:stability} shows that our student structure can learn the visual representation of different SOTA MIM methods, indicating the effectiveness of the sparse convolution and the UNet structure. Meanwhile, when the backbone is changed, our method can be applied to different depths and maintain a better accuracy that outperforms the supervised baselines.

\paragraph{Sparse convolution, UNet structure, and similarity distribution.} These are the three main features of our design to enable the student to mimic the visual representation of the teacher. To speed up the ablation study, we used 25\% data of ImageNet 1K and cached the distilled features or similarities (if needed). By replacing the sparse convolution with the traditional convolution, we find a degradation by 0.8\%, and by substituting UNet with same-depth CNN, a reduction of 0.5\% appears. Moreover, if we minimize the MSE loss for outputs solely, the accuracy will have a 0.9\% down-gradation. The pairwise combinations of ablation also bring accuracy degradation as shown in Table \ref{table:ablation}. Note that sparse convolution brings the most significant improvement, indicating the importance of inner relations. Therefore, all the strategies are effective.

\begin{table}[htbp]
  \caption{\textbf{H-GKD ablation experiments.} explores the effectiveness of our approaches. The default settings for our methods are in \colorbox{lightgray}{gray}. We choose CAE-L as our teacher and sparse ResNet-50 as our student}.
  \label{table:ablation}
  \centering
  \resizebox{0.75\linewidth}{!}{
    \begin{tabular}{|l|l|l|l|l|l|}
    \toprule
        methods & sparse cnn & Unet & similarity & epoch & TOP-1 Acc. \\ 
    \midrule    
        \colorbox{lightgray}{our} & \faCheck & \faCheck & \faCheck & 100 & 71.84 \\ 

        w/o Unet & \faCheck & \faTimes & \faCheck & 100 & 71.32 \\ 
        
        w/o sparse & \faTimes & \faCheck & \faCheck & 100 & 71.04 \\ 
        
        w/o similarity & \faCheck & \faCheck & \faTimes & 100 & 70.92 \\
        
        only w similarity & \faTimes & \faTimes & \faCheck & 100 & 69.77 \\ 
        
        only w sparse & \faCheck & \faTimes & \faTimes & 100 & 69.41 \\ 
        
        only w Unet & \faTimes & \faCheck & \faTimes & 100 & 69.12 \\ 
    \bottomrule
    \end{tabular}
    }
\end{table}

\section{Conclusion}

In this paper, we describe  a heterogeneous knowledge distillation method based on MIM, aiming at leveraging  the powerful ability to learn knowledge from reconstructing masked unlabeled data. We achieve a simple distillation method to train a sparse CNN-based student network based on a sizeable heterogeneous model. Extensive experiments show that our method outperforms several SOTA-supervised distillation methods. This provides ample space for further exploration in future work, such as low-level vision and AIGC.



\printbibliography

\clearpage
\appendix
\nolinenumbers

\section*{Appendix}
\label{appendix}

\subsection*{Psudo-implementation}
\label{appendix:psudo-imp}

The psudo-code of our H-GKD implementation is provided:

\begin{lstlisting}

'''
    S: student model
    T: teacher model
    Queue: the similarity queue shared with teacher and student
'''

# init the teacher model with generative self-supervised methods (such as MAE/CAE/BEiT)
T = init_teacher(args) 
# init the student model with sparse convolution
S = init_student(args) 

# load the pretrained teacher weights and states
T = load(pre_trained_T)

# teacher model is frozen
T.eval()

# enumerate
for imgs in data_loader:
    # functions to images, like scaling, flipping, jittering
    imgs = Handle_img(imgs) 
    
    with torch.no_grad():
        # frozen teacher and output norm teacher features
        T_features = T(imgs) 

    S_features = S(imgs)# output norm student features

    # MSE loss for teacher features and student features
    L_feat = L_f(S_features, T_features) 

    # teacher features en-queue first and calculate first
    Q.enqueue(T_features)
    Q.cal_similarity()
    # then for student
    Q.enqueue(S_features)
    Q.cal_similarity()

    # compute the similarity loss through the updated Q
    L_Pear = Pearson(Q)

    Q.dequeue() # de-queue the last instances

    (L_feat+L_Pear).backward() # update gradient
    
\end{lstlisting}

\end{document}